# A Compliant, Underactuated Hand for Robust Manipulation

Lael U. Odhner[1]*, Leif P. Jentoft[2], Mark R. Claffee[3], Nicholas Corson[3], Yaroslav Tenzer[2], Raymond R. Ma[1], Martin Buehler[3], Robert Kohout[3], Robert D. Howe[2], Aaron M. Dollar[1]

## Abstract

This paper introduces the i-HY Hand, an underactuated hand driven by 5 actuators that is capable of performing a wide range of grasping and in-hand manipulation tasks. This hand was designed to address the need for a durable, inexpensive, moderately dexterous hand suitable for use on mobile robots. The primary focus of this paper will be on the novel minimalistic design of i-HY, which was developed by choosing a set of target tasks around which the design of the hand was optimized. Particular emphasis is placed on the development of underactuated fingers that are capable of both firm power grasps and low-stiffness fingertip grasps using only the passive mechanics of the finger mechanism. Experimental results demonstrate successful grasping of a wide range of target objects, the stability of fingertip grasping, as well as the ability to adjust the force exerted on grasped objects using the passive finger mechanics.

## 1 Introduction

Capable robot hands are important for a wide range of applications, such as industrial warehouse operation, household chores, and disaster relief. To address these needs, hands must be inexpensive, compact, and robust, and must be capable of performing simple grasping and manipulation tasks, such as robust precision and power grasps, in-hand grasp transitions, and basic tool use tasks (operating switches, triggers, pliers, etc). At the present time, most commercially available robot hands are either single-actuator parallel jaws [1-3], or simplified multi-fingered hands optimized for one or two grasping configurations [4-7]. Extensive research has been dedicated to the development of highly articulated, general-purpose robotic hands and prosthetics [8-23], but in spite of continuing advances, most of these hands have not developed broad user bases beyond the groups or collaborations in which they were developed.

Numerous explanations can be found for this gap between research and practice; limitations in actuators, tactile sensors, fabrication methods, and control software are all barriers to progress. However, the assumptions commonly made during the process of robot hand design may also be to blame. General-purpose robot hands are often designed by starting with a high-level organizing principle that dictates the hand's major features. For example, many hands are strictly anthropomorphic, under the hypothesis that a robotic reproduction of a human hand will be capable of performing human tasks. Another frequently-used organizing principle is kinematic/force synthesis, particularly the requirement that a hand should be capable of applying arbitrary motions and forces to a grasped object. This leads naturally to strong mathematical design requirements for finger actuation and kinematics, often rooted in eigenvalue analysis of the fingertip Jacobians throughout the hand workspace [24-26].

---

[1] Dept. of Mechanical Engineering and Materials Science, Yale University, New Haven CT.
  E-mail: {lael.odhner, raymond.ma, aaron.dollar}@yale.edu
* corresponding author
[2] School of Engineering and Applied Sciences, Harvard University, Cambridge, MA.
  E-mail: {ljentoft, ytenzer, howe}@seas.harvard.edu
[3] iRobot Corporation, Bedford, MA.
  E-mail: {mclaffee, ncorson, rkohout}@irobot.com,mxbuehler@gmail.com

Top-down principles are appealing because they provide clear sufficient conditions for the generality of any robot hand, but this generality is a two-edged sword. Because they focus on guaranteeing that a hand can perform *all possible grasping and manipulation tasks*, principles such as strict anthropomorphism and kinematic/force synthesis lend little insight into *which tasks matter most, and which design choices strongly affect the performance of these tasks*. As a consequence, general-purpose hands often include a dozen or more actuators and elaborate transmission and sensing systems, resulting in hands that are heavy, expensive, and fragile. Problems of control are complicated by the vast number of sensor signals and actuators that must be coordinated in order to perform even a simple task. The complexity, fragility and expense of these hands also discourage unconstrained experimental testing.

In this paper, we introduce the iRobot-Harvard-Yale Hand (i-HY), a moderate-complexity robot hand with five actuators, capable of performing a variety of tasks, including power and fingertip grasping and simple in-hand manipulation (Fig. 1). Rather than pursuing a top-down design strategy, the development of i-HY followed a bottom-up process, beginning with the analysis of the range of tasks that the hand was required to perform, and culminating in a minimalistic design capable of performing these tasks. Many robot hands, often described as underactuated or passively adaptive, have been designed within this paradigm for the primary purpose of reliably performing just one or two basic grasps [27-32]. For example, the SDM Hand, upon which the design of i-HY was based, was developed for acquiring an enveloping grasp on objects of uncertain shape and position [31]. Mechanical features, such as passive compliant joints and differentially actuated tendons, ensured that grasps were stable even in the absence of sensory feedback to the hand. This new hand demonstrates that bottom-up design principles can be extended to hands of intermediate complexity. The i-HY Hand also incorporates the passive mechanical behavior and robustness for use in unstructured environments that made the SDM hand successful.

We begin with an analysis of the range of tasks that the hand is designed to perform, followed by presentation of the hand design itself. Section 3 shows how the design of the i-HY fingers embody the mechanical intelligence needed to perform each of the tasks using passive mechanics in place of elaborate sensing and control. We conclude in Section 4 with examples linking specific tasks to the mechanical features incorporated into the design of the hand.

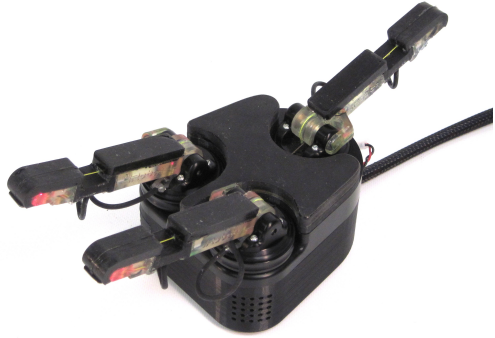

Fig. 1.i-HY is an underactuated hand capable of performing a wide range of tasks, including fingertip grasping and manipulation.

## 2    Task Analysis and Hand Design

We define 'bottom up hand design' to mean defining the hand specification in terms of tasks the hand must perform, rather than high-level mathematical sufficient conditions. This is not a strictly empirical process; mathematical rules can (and should) be applied to the design of the hand, but these rules should arise from analysis of the primitive operations required for the execution of each task, and should be biased to provide a minimalistic set of design constraints. These tasks are then distilled into grasping and manipulation primitives, and the hand layout, actuation, and finger structure are designed to meet them. This section describes the tasks that i-HY was designed to perform, and overviews the major structure of the robot hand design. The subsequent section will then discuss some of the key analytical questions addressed in the design of i-HY, particularly in the development of multi-purpose underactuated fingers.

### 2.1    Task Requirements

To apply a task-centric design approach, the first step is to define a representative set of tasks that the hand should be able to perform. These can be tailored to a specific setting (e.g. an industrial application), but they can also instead be selected to span a wide range of useful behavior, based by careful selection of representative tasks that span a range of capabilities. Existing studies on Activities of Daily Living may be helpful in this [33, 34]. The i-HY Hand was developed through the hardware track of the DARPA Autonomous Robotic Manipulation program (ARM-H), a broad effort to reduce the cost and increase the capabilities of hands for mobile manipulation. The requirements for ARM-H were delivered to participating teams as a list of challenge tasks:

- Pick up a key and putting it into a lock, then unlock and open a door
- Open a zipper on a backpack and remove the contents
- Pick up a pair of wire cutters and cut a wire
- Pick up and write with a whiteboard marker
- Grasp a radio handset and activate the push-to-talk button
- Grasp a drill and use it to drill a hole

- Grasp and turn on a flashlight from an unknown initial pose
- Grasp a hammer in a fashion suitable for use from an unknown pose
- Grasp and move a heavy, unknown object such as a rock or cinder block

Interestingly, this list overlaps with evaluation criteria used for prosthetic hands [35]; both focus on access (locks, doors or zippers) and basic tool use (hammers, drills, wire cutters, flashlights, etc.).Each evaluation task was analyzed to determine possible strategies for execution using as few actuators as possible. Bench-level prototypes, constructed using Shape Deposition Manufacturing (SDM) and 3D printing, were used to evaluate these strategies, and simplified analytical models were used to identify the critical parameters of each task.

This analysis culminated in a set of primitive grasping and manipulation operations deemed necessary for performing the manipulation tasks. Figure 2 illustrates the basic grasp types for which the hand was designed. The core function of any hand is typically the ability to execute enveloping grasps such as the power grasps shown at right. These are important for firmly attaching the grasped object to the end of the arm, and were needed at some point in most of the challenge tasks. Fingertip grasps, shown at left in Fig. 2, were needed for picking up small objects, but also for acquiring initial grasps on larger objects, especially if these were acquired from the surface of a table where a power grasp could not be directly acquired. Manipulation primitives necessary to perform the challenge tasks were also identified, and are depicted in Fig. 3. The most useful in-hand manipulations were found to be shifting from one grasp type to another, such as transitioning from a pinch grasp to a power grasp (shown top left in Fig. 3), or in adjusting the orientation of the object relative to the finger contact points. For instance, picking up a key and putting it into a lock entails first reorienting the key into a pinched configuration between the fingers (bottom left), then rotating the key so that the blade is aligned with the axis of the lock (top right), as in [36]. Button-pushing tasks are important to using many common tools such as drills, so the ability to perform a reversible single-fingertip squeezing motion (bottom right) was also required.

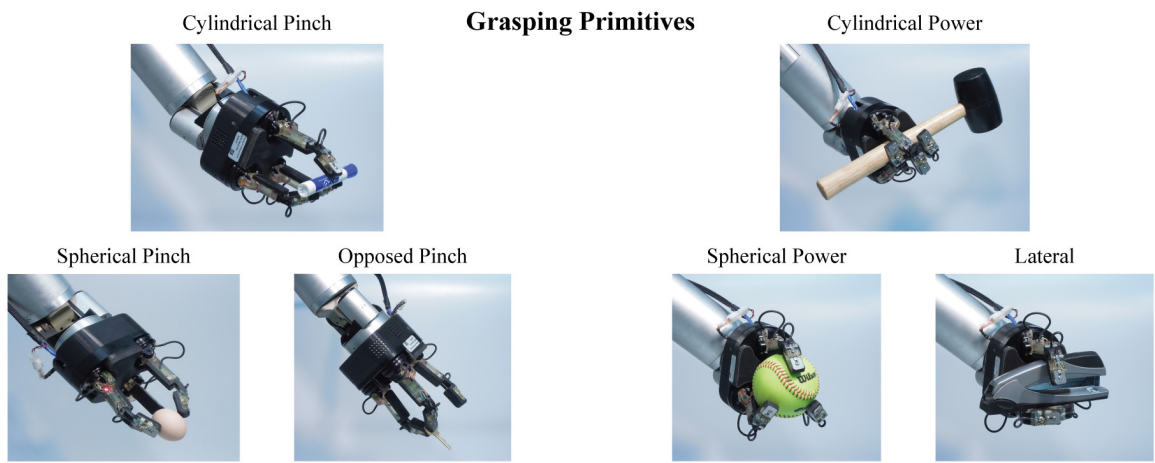

Fig. 2.The set of passive adaptive grasps that i-HY was designed to perform is shown above, including cylindrical, spherical and opposed pinch grasps, as well as cylindrical and spherical power grasps. The lateral grasp (bottom right) is performed by rotating two fingers so that they are normal to the plane of finger motion, providing a stiff surface against which the third finger can push.

**Manipulation Primitives**

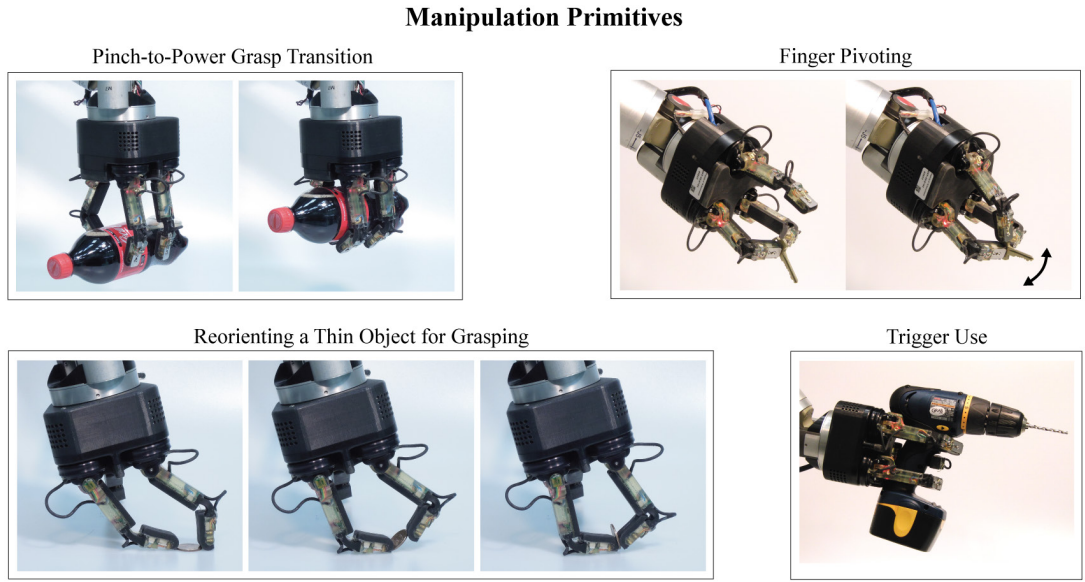

Fig. 3.Several key manipulation primitives were identified from the list of required tasks. Among these, the most important were the ability to transfer an object from a pinch grasp securely into a power grasp (top left), the process of rotating a thin object such as a key to acquire a power grasp (bottom left),the ability to pivot an object grasped between two fingers using the third (top right), and the usage of simple triggers and buttons (bottom right).

## 2.2   Hand Structure and Actuation

The goal in the design of i-HY was to use the simplest possible hand to embody all of the manipulation primitives. In this spirit, a basic three-fingered hand design was chosen, primarily because three is the smallest number of fingers that can be used for all of the required grasp types shown in Fig. 2, especially the cylindrical and spherical power grasps. The fingers were mounted to the palm in a triangular pattern, depicted in Fig. 4, so that a pair of fingers were placed in opposition to a third, hereafter called the thumb. The finger pair was actuated to adduct and abduct in a coupled fashion, so the fingers could be rotated between an opposed configuration, a spherical configuration meeting at a point in the center of the hand workspace, and an interlaced configuration suitable for grasping cylindrical objects. In this way, the finger/palm layout is similar to the Barrett Hand [4], although it accomplishes the movements in a different way.

The actuators were chosen to minimize cost and control complexity. Each of the three fingers is driven closed using a single flexor tendon running the length of the finger so that the fingers move along some compliant minimum-energy trajectory when not in contact with any objects [29, 31]. The actuators are Maxon brushless DC motors (EC-20, Maxon Motor AG, Sachseln, Switzerland), each connected to a tendon spool through a non-back drivable worm gear. These high-impedance actuators minimize both cost and power consumption, particularly to avoid overheating while holding static grasps. Because the worm gears are essentially non-backdriveable, almost no current is needed to operate the actuators in steady state, even in the presence of a significant load on the fingers.

A fifth actuator powers an extensor tendon on the proximal joint of the thumb. This actuator allows the link angle of the thumb proximal and distal joints to be set independently, and is particularly useful for tasks in which the tip of the thumb needs to move arbitrarily in the plane, such as the finger pivoting shown at lower left in Fig. 3. Because the extensor tendon has no antagonistic return tendon, it can be slackened so that the three fingers function identically. The entire hand weighs 1.35 kg, and the palm assembly housing the actuators fits into a package only 82 mm from the wrist to the palm surface, comparable to available commercial robot hands [4,5]. The total of five actuators places the hand in the middle of the range of the robotic hands available for research and prosthetic applications, well below most general-purpose research hands but above most underactuated grippers [37].

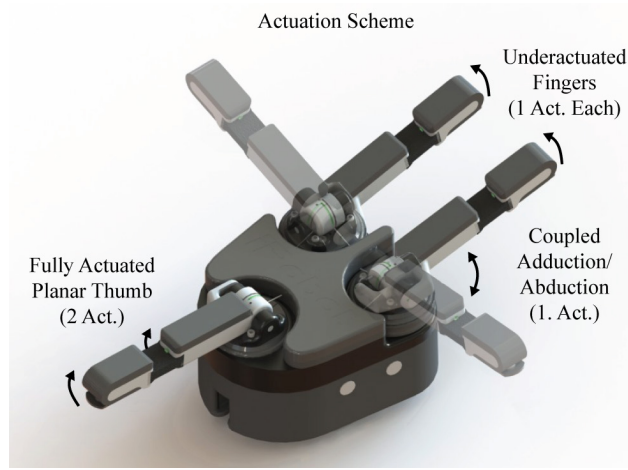

Fig. 4. A total of 5 actuators are used in i-HY: Each finger has an identical actuator controlling finger flexion. The thumb has an additional tendon so that the position of the thumb in the plane can be controlled. The two underactuated fingers have a coupled adduction/abduction motion to switch from cylindrical, spherical and opposed grasps.

## 2.3   Finger Design

The i-HY fingers, depicted in Fig. 5, utilize a two-link modular design similar to its precursor, the SDM Hand. The proximal pin joint connecting the finger to the hand is mounted on a circular magnetic base (also shown in Fig. 5), which serves as a modular attachment point, and also as a breakaway coupling to ensure that fingers will be minimally damaged in an unplanned collision. Spring-loaded "pogo pin" electrical contacts further simplify the modular fingers so that replacing a finger can be accomplished by simply snapping the finger in place, then attaching the flexor tendon. A tendon attachment point for the antagonistic tendon used on the thumb is included at the base of the proximal link on all fingers, so that any finger can be interchangeably used in any position on the hand.

The fingers were constructed using a casting and overmolding process based on Shape Deposition Manufacturing [38-40], but streamlined for higher-volume production by the elimination of intermediate machining steps. The various internal components of each finger (circuit boards, sensors, wiring and cable guides) were inserted into a mold cavity. Elastomeric parts, such as finger pads and flexure joints, were pre-molded in separate cavities and inserted along with the other components. A glass-filled epoxy resin was then injected into the mold, forming a single monolithic part. Figure 5 depicts a cross-section of the i-HY finger. Like the SDM Hand, the distal joint of each finger is an elastomer flexure. A more conventional pin joint with an in-molded bushing was chosen for the proximal finger joint. Because the fingers are sealed, they can resist water, dirt and impact.

One of the more important features of the i-HY fingers is their high compliance, especially at the distal flexure. The distal flexure hinge admits out-of-plane motion, illustrated in Fig. 6. The proximal pin joint includes a torsion spring, which provides the proximal joint with some elasticity. This compliance serves several purposes: first, because the fingers do not have extensor tendons, the joint elasticity alone extends the fingers when the flexor tendons are relaxed. This is particularly useful when operating the hand in an unknown environment were collision with obstacles is likely, so that fingers merely deform in response to unplanned contact. The torsional compliance at the distal flexure joint provides a similar robustness to the fingertips for out-of-plane contact. The second purpose served by passive finger compliance is passive adaptation to the shape of the object grasped, which removes the need to detect and react to small variations in surface geometry.

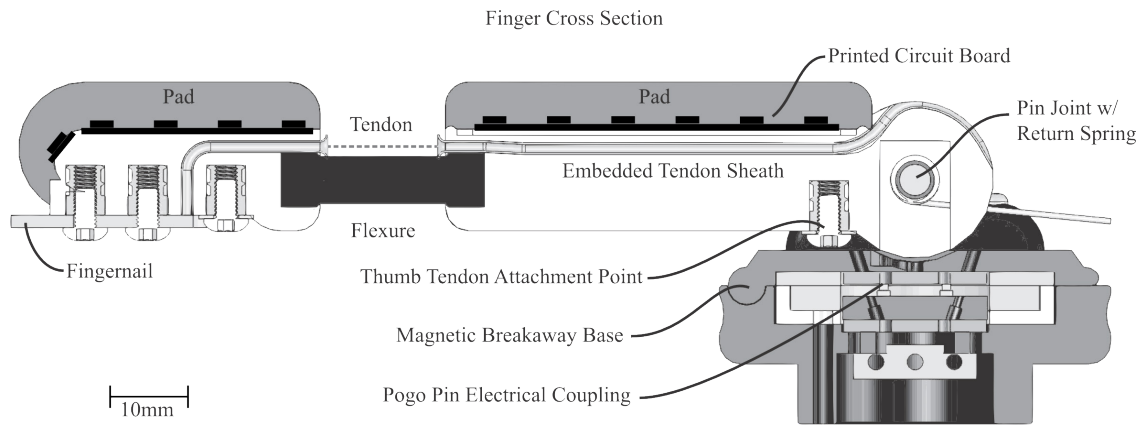

Figure 5. Cross-section view of i-HY finger design, showing the components embedded in the molded monolithic finger.

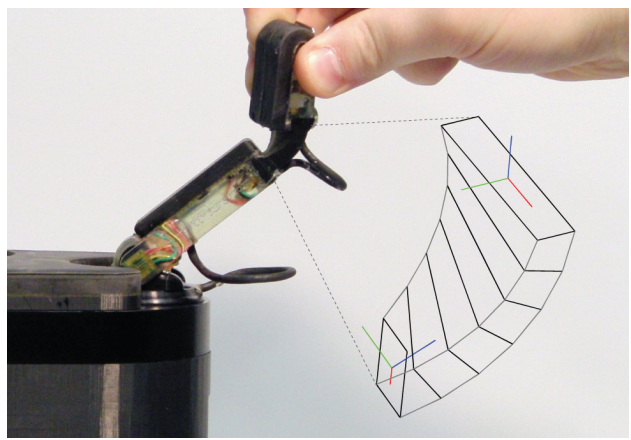

Fig. 6. The compliant flexure joints on the i-HY fingers allow three-dimensional fingertip motion.

## 2.4   Sensing

The i-HY sensor system (Fig. 7), consists of tactile arrays covering the fingers and palm, flexure deformation sensors in each distal finger joint, magnetic encoders at each proximal finger joint, and accelerometers in the distal finger links. Philosophically, the simplest sensor system delivering the necessary information for grasping was sought, with particular focus on measuring the hand forward kinematics and contact detection/localization. Wherever possible, commercial, off-the-shelf parts were used to maximize the reliability and minimize the cost of the hand. Two of the sensors developed for the hand are novel: sensors for detecting the three-dimensional deformation of the distal flexure joints (Fig. 6), and inexpensive MEMS tactile arrays for localizing the contact of objects with the hand.

Flexure deformation gives important information about contact detection [41], object localization, and measuring forces [42]. However, there is no prior work on large-deformation, multi-DOF flexure sensors, so a new sensor was created. Rotation around two axes are measured at each end of the joint, using four modules that measure local angle each using an optical fiber shining onto a pair of surface-mounted phototransistors. The rotation rate is integrated along the length of the joint using an approach based on [43]. This makes it possible to evaluate joint flexion, twist, combination of these, and shear. It can be combined with tendon excursion to reduce sensitivity to higher-order deformation modes not observed by the four-point angular measurements.

The tactile arrays used commercial off-the-shelf MEMS barometers (MPL115A2, Freescale Semiconductor, Austin, Texas) to provide tactile sensing with 10mN sensitivity and 4.9N range by casting the sensor circuit boards inside the molds for the fingers and palm, so that the rubber pads are firmly bonded to the sensors [44]. Embedding the barometers, soldered onto standard surface-mount boards, solves the systems integration problem that has hampered use of tactile sensors in hand designs [45, 46]. These are laid out in a 2x6 array on proximal links, a 2x5 array on distal links (with two wrapped around the fingertip), and a pattern of 48 on the palm concentrated in areas where contact is most likely, such as the edges of the palm (Fig. 7). Presently, tactile data on all sensors is available at 50 Hz through the data bus running along each finger into the palm.

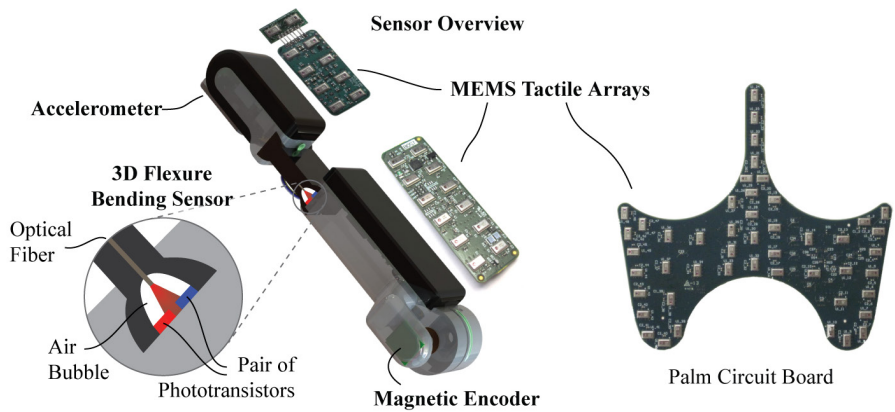

Fig. 7. Each finger contains joint angle sensing via a magnetic encoder, optical flexure bending sensor, and an accelerometer. Tactile sensing is provided by an array of MEMS pressure sensors.

# 3    Compliant Underactuated Fingers

A well-defined set of grasping primitives makes it possible to involve passive mechanics and compliance in grasping control, rather than relegating all questions of stable task performance to the software domain. This has already shown significant advantages for power grasping, as demonstrated by the success of underactuated grippers [29-31]. This decreases the cost and complexity of the robot, and reduces the sensory information required to perform a task. It also increases the robustness of the hand to unexpected collisions with the surrounding environment, particularly when such obstacles are usually close to the target object. This section explains how the design of the underactuated i-HY fingers was tuned to perform basic compliant power grasps like an underactuated gripper, as well as new and more dexterous fingertip grasping and manipulation operations.

## 3.1    Designing Compliant Fingers for Robust Power Grasps

Although compliance provides many advantages [47], it is not easy to build good impedance-controlled actuators. Small, inexpensive motors tend to be stiff when geared down sufficiently to provide high forces. Fixing this in software necessitates high-bandwidth, high-fidelity sensing and control loops which are expensive to implement. Series-elastic actuators (SEAs) instead address the problem mechanically by placing a spring on the output of a position-controlled motor [48]. This allows low-impedance control, and the mechanical system responds to high-bandwidth loads without requiring high-bandwidth control loops. Using compliant flexure joints and underactuation, it is possible to design fingers that have many of the advantages of SEAs, including error tolerance, reduced sensor load, and passive adaptation to object variation. The process of obtaining a power grasp is illustrated in Fig. 8. First, the fingers sweep in around the object. In order to maximize the chance of enveloping an object, the fingers must remain straight during this phase of motion. Once the fingers have contacted the object, the distal finger links must then cage around the object in order to complete the grasp. A variety of mechanisms have been used to achieve this behavior with a single actuator per finger. These include brakes in the joints, which must be switched off when contact is detected [49-52], as well as friction clutches that move only when the force on the finger tendon exceeds some threshold, as in the Barrett Hand and others [28, 53, 54]. However, both of these approaches introduce undesirable behaviors into the hand – brakes increase the mechanical complexity of the finger, and require active control. Although a clutch is mechanically simpler, it cannot be moved reversibly. Once the fingers have begun caging, only a complete extension of the finger will reset the clutch. Instead of relying on these mechanisms, an underactuated elastic design was used, similar to the SARAH, GRASPAR and SDM Hands [29-31]. Because these fingers have fewer actuators than phalanges, the actuator position or angle only partially constrains the configuration of the finger. The free motion of the finger can only be fully determined when considering the conditions for equilibrium in the finger [31, 55]. With the SDM Hand and i-HY, the proper motion is achieved by designing the distal finger joints to be much stiffer than the proximal joints, as analyzed in [31] and in [56].

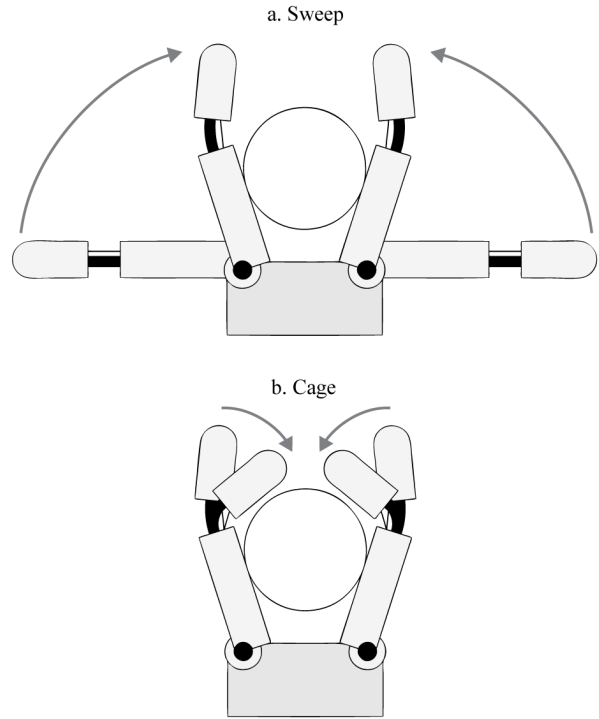

Fig. 8.The process of acquiring a power grasp can be divided into approximately two phases: the sweeping phase, in which the proximal finger links are brought into contact with the object, and the caging phase, in which the distal links flex to encircle the grasped object.

The compliant behavior of the i-HY finger can be understood by considering the simplified model shown at the bottom of Fig. 9, consisting of a spring-loaded 4 bar linkage in which the tendon serves as the fourth link running parallel to the proximal joint. A more accurate model for the i-HY finger, also shown in Fig. 9, would include several higher bending modes of the distal flexure joint as described in [43], and is derived in detail in [56]. For both of these models, the motion of the fingers during the sweeping phase of grasping can be modeled by treating the length of the flexor tendon as a rigid constraint on the system, then by minimizing the elastic energy in the two joints to find the equilibrium finger configuration as the finger closes. Once a single contact has been made and the fingers start caging, the underactuated finger acts as a differential transmission, exerting a torque on each of the finger links proportional to the moment arm of the tendon at each joint. Thus, the force on the proximal finger links will increase gradually as torque builds up on the elastic distal joint. Once the hand has fully closed, the fingers are no longer underactuated in the sense that their motion is fully constrained by contact with the grasped object. Due to higher-order elastic deformations, such as the rubber finger pads and the internal deformation modes in the flexure joint, the power grasp will not be infinitely stiff, but it will be sufficient to hold even heavy objects.

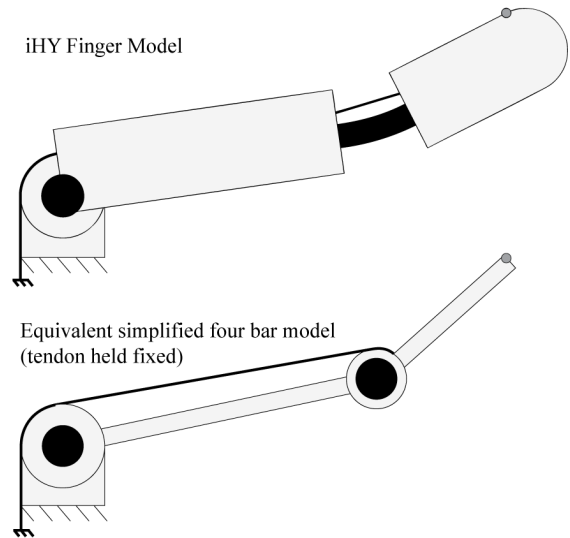

Fig. 9. The i-HY fingers were modeled as two-link underactuated mechanisms by constraining the total tendon length on the proximal and distal joints, while allowing the joints to deform elastically. This model was used to approximate finger compliance on the proximal and distal links.

In order to demonstrate the hand behavior during power grasping, the contact forces on a grasped object were measured in a simple experiment that captured the salient features of the i-HY finger mechanics. This was accomplished by constructing a cylindrical object split down the middle, having a load cell embedded in it as depicted in Fig. 10. The underactuated fingers were moved into an opposed, planar configuration, and closed slowly on the grasped object. The load cell measured the force exerted between the two fingers in the direction of the split, which was oriented such that it was symmetric with the fingers, as shown in the figure. The results, plotted in Fig. 11, confirm that the internal forces on the object in the direction of the split remain low (< 3 N) until the caging process is complete. After this point, further excursion of the tendons causes a linear increase in internal force, to the point measured.

The underactuated behavior of the finger can be clearly seen in the changing slope of the force-excursion curve as the fingers close. In the caging phase, the mechanism is underconstrained, and consequently force builds up slowly due to bending of the flexure. After the distal links have also made contact, the whole hand mechanism is stiffened by the added constraints, so much larger forces can be exerted on the object. In addition, because the caging motion is reversible except for a small amount of viscoelastic hysteresis in the flexure joint, tasks in which trigger-like motions are needed (operating a power drill, for example, or depressing a button on a radio handset) are accomplished without specialized mechanisms or control functions. Finally, it is important to remember that although the forces measured in this experiment are small, the fingers can resist much larger forces than they can actively exert due to the non-backdriveable tendon actuators. For example, the video attachment to this paper depicts i-HY picking up a 22kg weight using a cylindrical power grasp.

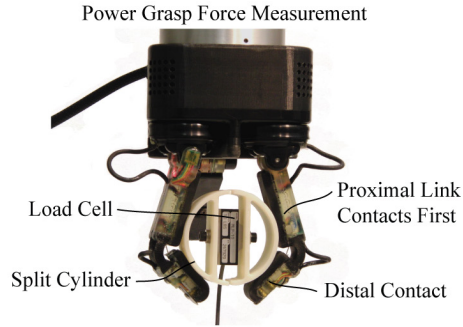

Fig. 10. An apparatus for measuring the internal force on an object in a power grasp configuration.

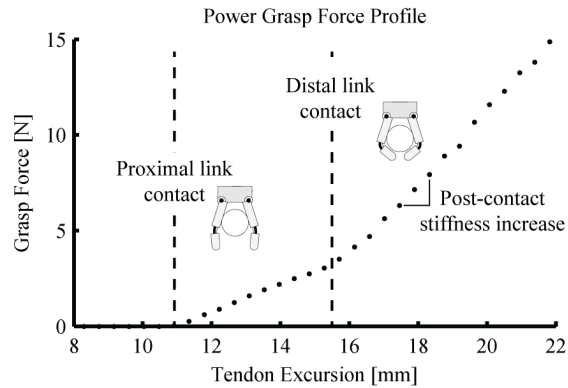

Fig. 11. Power grasp force on a 65 mm wide object as tendons are tightened. The "knee" in the curve corresponds to the point at which the distal links make contact with the object, stiffening the whole hand.

## 3.2    Extending Compliance to Stable Fingertip Grasps

Like power grasping, fingertip grasping and manipulation tasks also place requirements on finger compliance. In order to apply predictably low forces and adapt to the shape of grasped objects, compliance should be high in the direction normal to fingertip contact, as shown in Fig. 12. However, they must also be somewhat stiff laterally to provide stable pinch grasps. Finally, fingers must be robust to being jammed into rigid surfaces because many fingertip grasps are required acquiring objects from rigid surfaces. In a conventional highly articulated hand, this compliance is often provided using active fingertip force or impedance control [57-59], or through the introduction of passive compliance in the finger actuators [60], but it is difficult to obtain this kind of high-fidelity performance in a compact hand. However, meeting these compliance requirements is doubly difficult in an underactuated finger because these fingers also have one or more uncontrollably compliant directions of motion due to their underconstrained design. For example, the four-bar linkage formed by the simplified model in Fig. 9 will move passively in one direction even if the tendon is locked. Attempts at obtaining stable fingertip grasps with underactuated fingers have typically involved the addition of special-purpose mechanisms for stabilizing the fingertips, such as active locking mechanisms to reduce underactuated degrees of freedom [49-52], specially designed fingertips to cup grasped objects [61], or strategically placed hard travel limits so that the fingers stiffen when moved into a pinching configuration [6, 29, 32].The key to the good fingertip grasping performance of i-HY is that this direction of unavoidable passive compliance due to underactuated is optimized so that it lies normal to the fingertip. This solves two problems at once: eliminating a source of instability, while causing the fingertips to act as a kind of virtual series elastic actuator, as shown below.

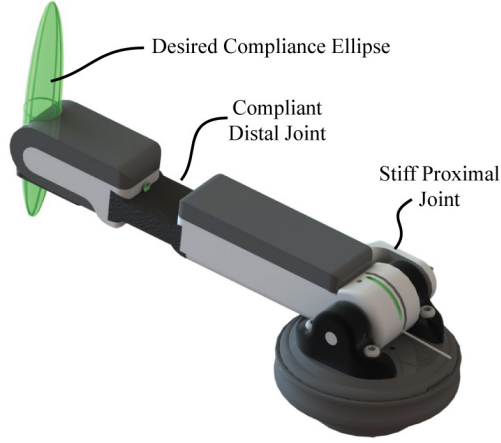

Fig. 12. The desired fingertip compliance ellipse is narrow in directions representing shear motion, and long in the direction representing normal motion. This means the finger will be able to move gently while still holding objects in a stiff grasp.

As Fig. 12 shows, the design goal of the i-HY fingertips was to place the direction of principal underactuated compliance normal to the tips of the fingers. This was accomplished by analyzing the variation of the finger compliance over the whole distal link. Based on the flexure-based finger model from [56], the compliance of the distal finger link was found by considering the variation in energy associated with the perturbation of a point on the distal link. This compliance was characterized at some point $x$ by a 6×6 matrix, $\mathbf{C}$, relating a small force $\delta f$ and a small torque $\delta \tau$ to a small body-frame displacement $\delta x$ and rotation $\delta \theta$:

$$\begin{bmatrix} \delta x \\ \delta \theta \end{bmatrix} = \mathbf{C} \begin{bmatrix} \delta f \\ \delta \tau \end{bmatrix} = \begin{bmatrix} \mathbf{C}_{xx} & \mathbf{C}_{x\theta} \\ \mathbf{C}_{x\theta}^T & \mathbf{C}_{\theta\theta} \end{bmatrix} \begin{bmatrix} \delta f \\ \delta \tau \end{bmatrix} \qquad (1)$$

The blocks $\mathbf{C}_{xx}$, $\mathbf{C}_{\theta\theta}$ and $\mathbf{C}_{x\theta}$ represent the Cartesian and torsional compliance, and the coupling between the two. At some other point $x' = x + d$ defined by a rigid-body translation, the Jacobian relating local body-frame motion at $x'$ to motion at $x$ is the adjoint operator, which can be written as a 6×6 matrix, $\mathbf{J}$,

$$\begin{bmatrix} \delta x' \\ \delta \theta' \end{bmatrix} = \mathbf{J} \begin{bmatrix} \delta x \\ \delta \theta \end{bmatrix} = \begin{bmatrix} I^{3\times3} & d_\times \\ 0 & I^{3\times3} \end{bmatrix} \begin{bmatrix} \delta x \\ \delta \theta \end{bmatrix} \qquad (2)$$

Here $I^{3\times3}$ is a 3×3 identity matrix and $d_\times$ is the 3×3 skew-symmetric matrix corresponding to the 3×1 vector, $d$. The small-force compliance at $x'$ can also be found using this Jacobian [57],

$$\mathbf{C}' = \mathbf{J}\mathbf{C}\mathbf{J}^T = \begin{bmatrix} I^{3\times3} & d_\times \\ 0 & I^{3\times3} \end{bmatrix} \begin{bmatrix} \mathbf{C}_{xx} & \mathbf{C}_{x\theta} \\ \mathbf{C}_{x\theta}^T & \mathbf{C}_{\theta\theta} \end{bmatrix} \begin{bmatrix} I^{3\times3} & 0 \\ d_\times^T & I^{3\times3} \end{bmatrix} \qquad (3)$$

Because (3) describes the compliance of any point on a rigid body given the compliance of a single point, it can be applied to map out the compliance ellipses (which look almost 1-dimensional due to the high stiffness in the orthogonal direction) along the length of the finger, as shown in Fig. 13.

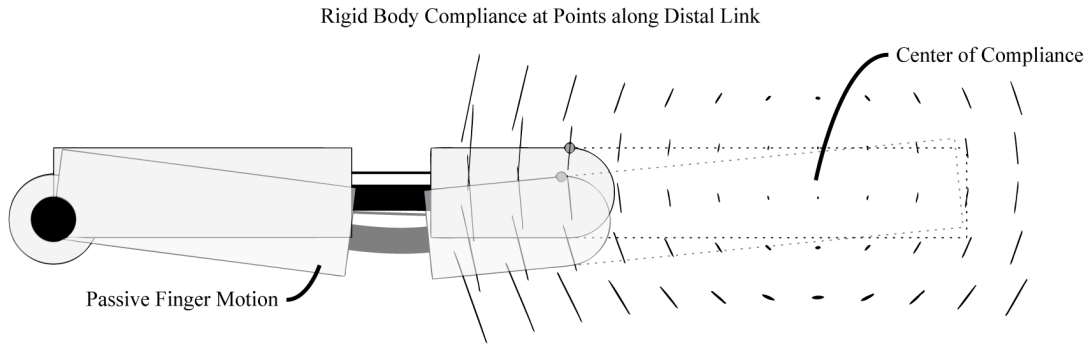

Fig. 13. The in-plane compliance of the finger is minimized at its center of compliance. The distal link design was shortened so that contact will occur at a point where there is compliance normal to the fingertip surface, and parallel to the direction of fingertip motion.

According to these results, the geometry of the distal link is very important to generating an appropriate compliance at the fingertip. Near the distal link's center of compliance, located at a point approximately 6 cm past the base of the distal link, the Cartesian compliance of the fingertip is very small. If the finger were long enough to make contact at this point, the fingertip would resist disturbance forces, but it would also move rigidly, and would not be at all backdriveable due to the stiff actuator tendon. Instead, the finger was shortened so that the finger exhibited significant compliance, principally in the direction normal to the fingertip. To demonstrate that this compliance property is preserved as the finger moves throughout its workspace, the equilibrium configuration of the finger was modeled along its closing trajectory, and the principal fingertip compliance was computed in each configuration (Fig. 14). The results of this analysis confirm that the principal direction of compliant motion remains more or less normal to the fingertip as the tendon is contracted. More importantly, the direction of motion of a point on the fingertip is always within approximately 30 degrees of the principal compliance, so that the finger will always deform if the finger is driven into the surface of an object. Thus, proper finger link geometry enables fingertip compliance in the direction of actuation.

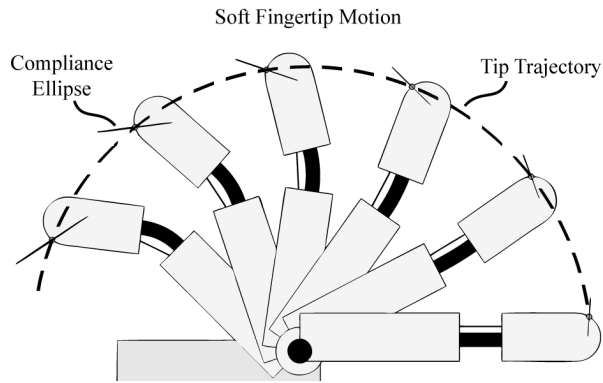

Fig. 14. The principal direction of compliance at a point on the tip of the finger, as the finger is moved through its range of motion. The principal direction of compliance is mostly aligned with the direction of motion.

The out-of-plane compliance of the fingers, due largely to the torsion of the distal joint flexure (Fig. 6) is also critical for producing stable fingertip grasps. The fingers on the SDM Hand were optimized for power grasps, and performed poorly when used for fingertip grasps. The same analysis used to determine in-plane compliance also explains out-of-plane compliance. One important implication of eqn. (3) is that the Cartesian compliance at any point on a rigid structure will vary as the square of the translational offset multiplied by the torsional compliance matrix,

$$\mathbf{C}'_{xx} = \mathbf{C}_{xx} + d_\times \mathbf{C}^T_{x\theta} + \mathbf{C}_{x\theta} d^T_\times + d_\times \mathbf{C}_{\theta\theta} d^T_\times \qquad (4)$$

The length-squared compliance term, $d_\times \mathbf{C}_{\theta\theta} d^T_\times$, means that the torsional compliance of the proximal joint will produce much more Cartesian compliance at the fingertip than the distal joint, because the moment arm from the fingertip to the axis of out-of-plane joint torsion is longer, as shown in Fig. 15. To compensate for this, the proximal joint, originally a flexure in the SDM Hand design, was replaced with a pin joint to increase torsional stiffness in out-of-plane motion. The distal joint, now the dominant source of torsional compliance, was adjusted experimentally to tune the out-of-plane fingertip compliance such that it provided useful compliance during grasp acquisition (e.g. conforming to a tabletop surface) but was stiff enough to enable stable fingertip grasps.

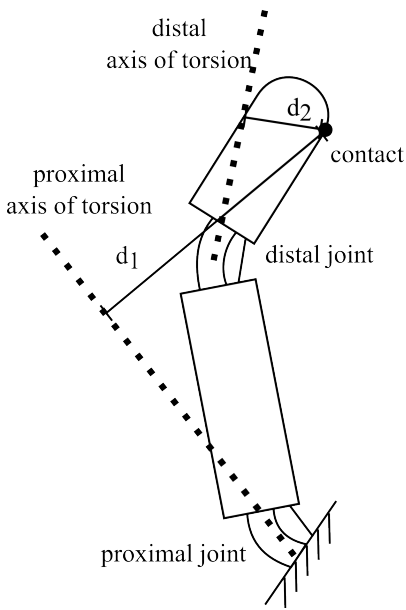

Fig. 15. With a flexure at the proximal joint, the proximal joint produces greater Cartesian compliance at the fingertip contact than the distal joint.

These results show that by tuning passive underactuated fingertip compliance, the same impedance properties desired by general-purpose robot fingers can be imparted to the i-HY fingers, using only a single flexor actuator per finger. Furthermore, because the compliance of the fingers is introduced through the underactuated design, neither a low-impedance servo motor nor a series elastic tendon actuator is required to obtain delicate fingertip grasping capabilities. Because of this, and because no additional components or control actions are needed to obtain good fingertip properties, the i-HY design also has significant advantages over many other approaches to fingertip grasping in underactuated hands.

The effect of compliance in fingertip grasps was measured using the same apparatus used to measure the force-excursion curve of a power grasp (Figs. 10 and 11). The instrumented cylindrical object was placed between the fingertips as shown in Fig. 16, and the force was measured as the flexor tendons on both fingers were contracted. The increase in contact force plotted in Fig. 17 as a function of the tendon excursion is gradual and linear, confirming that the fingertip behaves as a series elastic actuator. The linearity of the fingers is interrupted only at the point when the distal link reaches its travel limit, making contact with the proximal link. Because this adds an internal constraint on the fingers' motion, the force-excursion relationship stiffens past this point, and still remains predictable.

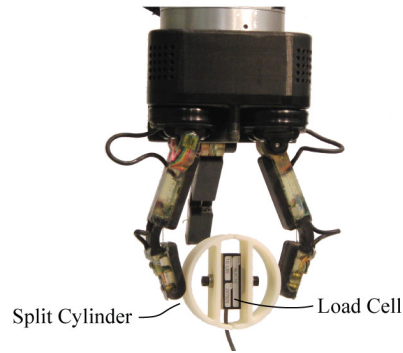

Fig. 16. The test apparatus used to measure the force-excursion curve on a 65 mm object in a pinch grasp.

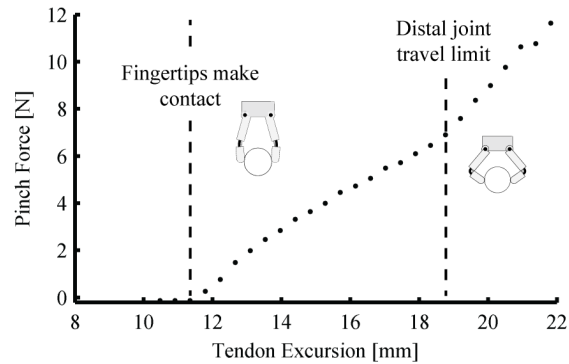

Fig. 17. The pinch force on a 65 mm wide object as tendons increases gradually and linearly as the tendons are contracted.

## 3.3 Summary

This section has shown how the minimalistic *fingers* of i-HY were designed to satisfy two task-based functional requirements: preserving the robust, adaptive power grasping capabilities of the SDM Hand and similar adaptive hands, while adding the ability to perform grasp with the fingertips as if they were driven by low-impedance actuators. Due to the unique impedance tuning of these fingers, no *ad hoc* mechanism was needed for fingertip manipulation, only the adjustment of the finger link lengths relative to the finger's center of compliance. Measured force-excursion curves showed that the effect of the passive elasticity on power grasping and fingertip tasks results in a predictable, mostly linear relationship between finger force and tendon excursion. In the final part of this paper, the whole- hand capabilities will be demonstrated on a series of real-world tasks.

# 4   Grasping Primitives in Task Space

This section shows how hand design features are leveraged for task execution. A series of real-world grasping and manipulation tasks was examined. The tasks include grasping spherical objects, operating tools with triggers, performing the pinch-to-tripod grasp transition, and acquiring power grasps on objects sitting on a flat surface. A few of these tasks are presented in the sections below to highlight the design features. A video from the demonstrations is included with this paper as an attachment.

## 4.1   Compliance

Compliance was identified as a critical design feature because it provides both structural robustness during collisions and finger adaptation to object geometry during grasping. Two separate tasks highlight this feature. The robustness is demonstrated by driving the fingers into the table, as in Fig 18. The fingers deform during the contact with the table without damage. The compliant fingers adaptation can be presented using the task of tightening up the lid of a bottle (see Fig 18). In this example the lid is held in a spherical pinch grasp and the center of rotation of the hand is misaligned with the axis of the bottle. Nevertheless, the finger compliance creates a mechanical coupling that makes it possible to keep a firm grasp of the lid while the lid is tightened/loosened.

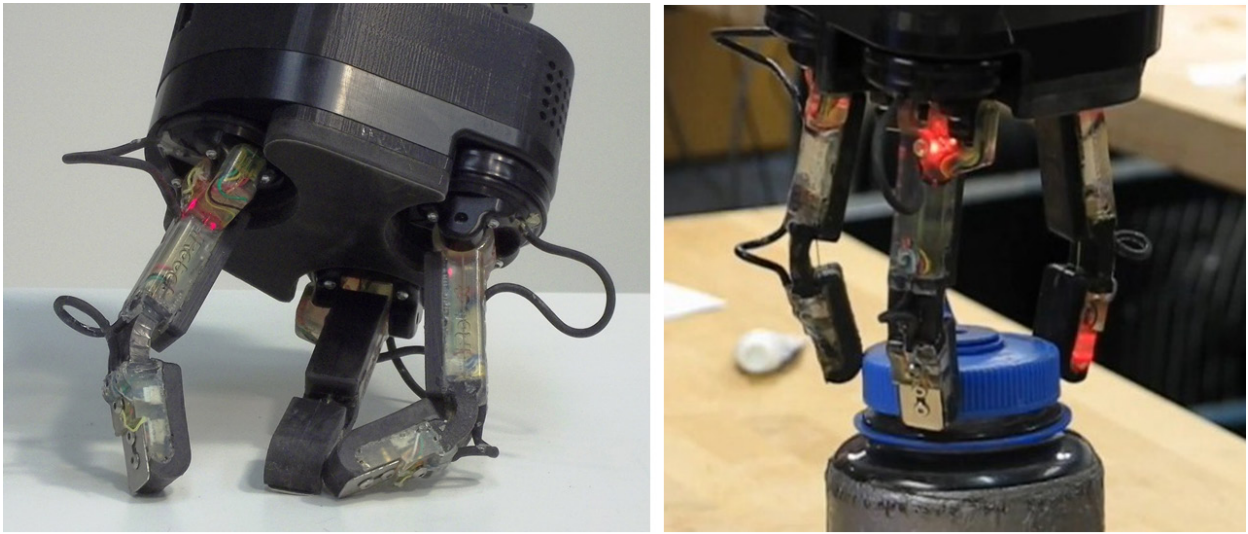

Fig. 18. Compliance during tasks; fingers deform during unintentional contact with a surface (left), and passive adjustment of the fingers during lid tightening task (right).

The combination of finger robustness and compliant adaptation simplifies automation during power grasp acquisition task (Fig 19). This task was implemented autonomously using a calibrated Microsoft Kinect sensor to acquire the centroid and principal axis of a series of objects placed on a table. Much like the SDM Hand [31], i-HY was capable of grasping each object simply by moving the center of the palm to the centroid of the object, and then closing the fingers on a pre-recorded trajectory. Fig. 19 (bottom) shows a set of objects grasped during this task. Of these, all but the ballpoint pen were grasped successfully greater than 19 of 20 attempts. The small size of the pen led to large errors in the point cloud, so pre-positioning of the hand was less accurate. For this object, 12 of 20 attempts were successful.

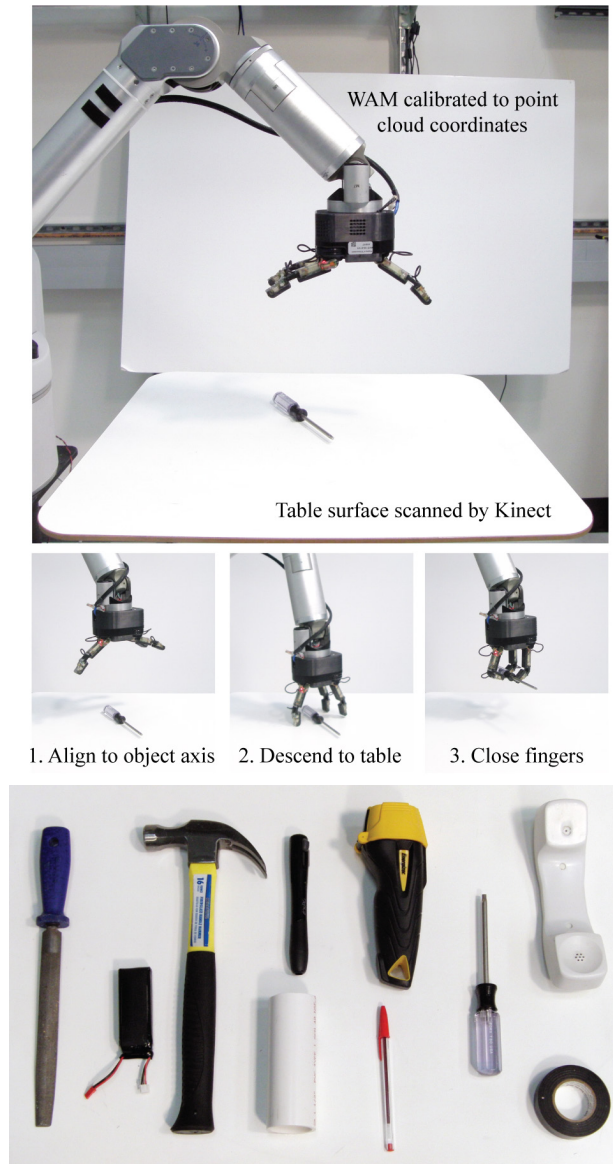

Fig. 19. (top and middle) Grasp acquisition task performed autonomously using an overhead Kinect sensor, moving the hand to the centroid of the object, aligning with the major axis, and grasping. (Bottom) Objects grasped during this experiment.

## 4.2 Variable Distal Joint Stiffness

The flexures in the distal joints of the fingers facilitate passive adaptation of the fingertips to the object geometry. Nevertheless, in some tasks the torsional stiffness of these flexures must be increased such that larger forces can be applied to the object. One of these tasks is turning a key in a lock, where large moments must be applied to key to perform the task. In this case the key is held in a pinch grasp and the torsional stiffness of the finger is increased by flexing the distal link towards its travel limit, greatly increasing the stiffness of the distal link (Fig 20).

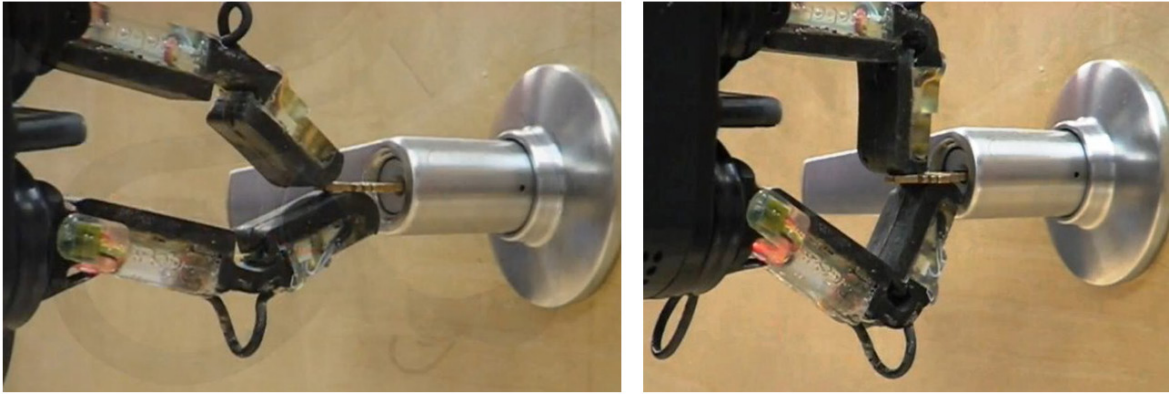

Fig. 20. Key insertion/turning task. The key is held in an opposed pinch grasp (left) and then the torsional stiffness of the flexure is increased by bringing the two links into contact at the distal joint.

## 4.3 Grasp Stability

Because underactuated fingers are free to deform elastically even when the actuators are locked, a completely rigid set of constraints on a grasped object cannot be obtained in a fingertip grasp. Instead, the object possesses some elastic potential energy well whose convexity (the stiffness of the grasp) determines the grasp stability [27, 61, 62]. If the grasp provided by the underactuated fingers is insufficiently stiff, then it is possible to strip an object from the grasp with a relatively small amount of work [63].

Examples of spherical pinch grasp stability for i-HY hand during task execution are illustrated in Fig.21. In one example a marker is held and external contact forces are exerted at the tip of the marker during a writing task; the fingers deform but the marker is held firmly. In another example an AA battery is grasped and is forcefully driven into the table. In this case the fingers deform much more prominently, but the object stays in the grasp.

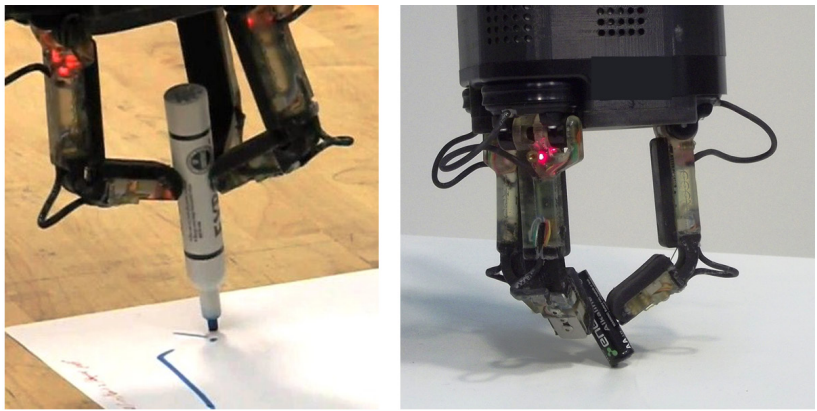

Fig. 21. Stability of the spherical pinch grasp; stability of the marker during writing (left), a grasped battery colliding with the table (right).

To demonstrate that the i-HY fingers acquire sufficiently stiff grasps to hold small and medium-sized objects in pinch grasps despite the compliant behavior of the fingertips, their stiffness was measured empirically with an apparatus shown in Fig. 22. A cylindrical test object 65 mm in diameter was attached to a 6 axis force/torque sensor (Gamma, by ATI Industrial Automation, Apex, NC, USA), and mounted in the headstock of a 3 axis milling machine, as shown in Fig. 21. The hand was held in the milling vise and commanded to grasp the test object. Disturbance displacements were applied to the object using the bed translation of the milling machine, and the resultant force was measured. This procedure was performed for a planar fingertip grasp of two opposed fingers, and a spherical fingertip grasp with all three fingers.

An example of the resulting data set is shown in Fig. 23. Some hysteresis was observed due to tendon friction and the viscoelasticity of the polymer pads and flexures, and this was dealt with by moving the test object in cycles to obtain an average of the stiffness over both directions of motion. Linear least squares estimation was used to fit the parameters of a symmetric stiffness matrix to the data for both the opposed and spherical fingertip grasps. The resulting estimated two-fingered grasp planar stiffness matrix, in units of N/mm, was:

$$\begin{bmatrix} K_{xx} & K_{xy} \\ K_{yx} & K_{yy} \end{bmatrix} = \begin{bmatrix} 0.445 & 0.0543 \\ 0.0543 & 0.409 \end{bmatrix} \tag{5}$$

The estimated three-dimensional Cartesian stiffness matrix of the spherical fingertip grasp was:

$$\begin{bmatrix} K_{xx} & K_{xy} & K_{xz} \\ K_{yx} & K_{yy} & K_{yz} \\ K_{zx} & K_{zy} & K_{zz} \end{bmatrix} = \begin{bmatrix} 0.569 & 0.0553 & 0.0323 \\ 0.0553 & 0.696 & 0.0755 \\ 0.0323 & 0.0755 & 0.809 \end{bmatrix} \tag{6}$$

In each of these results, the stiffness matrix was found to be well-conditioned, having no dominant diagonal terms and consequently no directions of atypically high or low compliance. The magnitude of the measured stiffnesses can be pictured by envisioning an apple weighing approximately 100 g, causing a deflection of approximately 2 mm for a stiffness of 0.5 N/mm. Therefore, although some deformation will be anticipated due to the weight of grasped objects or contact forces, any small or medium household object will not exceed the fingertip grasping capabilities of i-HY, despite their intrinsic compliance.

3 Axis Mill Experiment for Measuring Pinch Grasp Stiffness

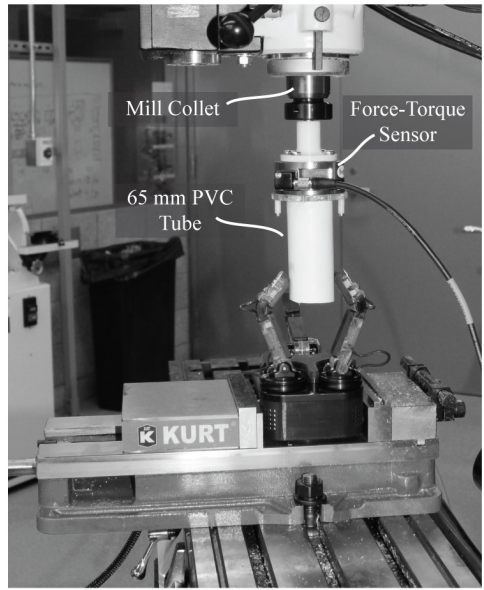

Fig. 22. Apparatus for measuring the compliance of planar and spherical pinch grasps. A 6 axis force-torque sensor held in a mill headstock is used to measure the force resulting from a rigid displacement in the grasp.

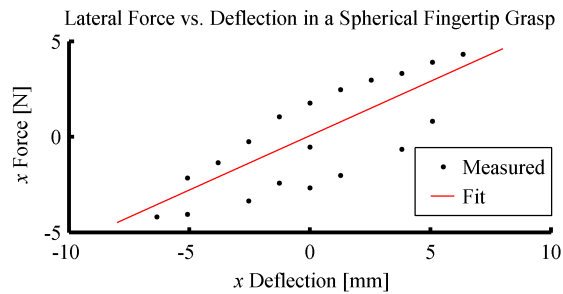

Fig. 23. A plot of the force versus deflection for a cyclic position disturbance on an object in a spherical fingertip grasp. The hysteresis observed is due to viscoelasticity in the elastomer pads and flexures, and friction in the tendon sheath.

## 4.4  In Hand Manipulation

Another example where the finger compliance proves to be useful is in-hand manipulation. We have shown in previously published work how a hand with the geometry of the planar opposed finger pair can be utilized to perform precision manipulation of an object grasped in a fingertip grasp [56]. The attached video also shows an example of three-dimensional fingertip manipulation, performing the task of picking up an AA battery from the table and inserting it into a flashlight. The battery can be picked up using an opposed pinch grasp, but can for insertion then be transitioned to a spherical pinch grasp, which is more robust and permits three dimensional manipulation. During the transition between the grasps the fingers adjust passively to the battery surface while maintaining grasping forces as the base of the fingers rotate into the new position. (Fig 24).

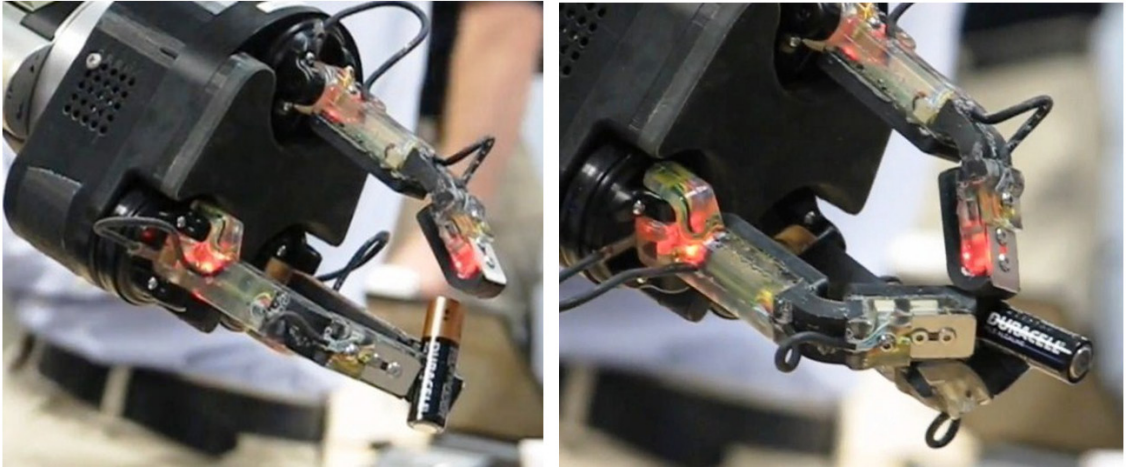

Fig. 24. Holding the battery in an opposed pinch grasp (left), and transitioning into the spherical pinch grasp (right) while maintaining grasp forces on the object.

# 5 Conclusions

The i-HY Hand is a medium-complexity manipulator with 5 actuators that is able to perform a wide range of tasks while maintaining a simple design that is robust, easy to control, and inexpensive. Through strategic use of compliance and underactuation, the hand is able to manipulate objects gently, and is also mechanically robust. A task-centric design approach extended the functionality of its precursor SDM Hand [22] which performed gripping but not manipulation operations. Much attention has been given to defining dexterity [46, 47] top-down in terms of Jacobian analysis or other hand-centric methods. However, at a task level, "dexterous is as dexterous does" and this bottom-up enables both appropriate choice of design trade-offs in terms of their end utility and comparison to hands designed using more traditional methods. The end result is a highly capable hand that can perform a wide range of tasks including lifting tools off flat surfaces and manipulating them into a power grasp, operating various tools with triggers such as drills and flashlights, and picking small flat objects such as keys off flat surfaces and moving them into a stable pinch grasp where they can be used.

# Acknowledgments

The authors are would like to thank Frank Hammond from the BioRobotics Laboratory at Harvard University, and Erik Steltz, Wes Huang, and Ben Axelrod at iRobot Corporation for their contributions and insights on the design of i-HY. Funding was provided under the DARPA Autonomous Robotic Manipulation program, hardware track (ARM-H), grant no. W91CRB-10-C-0141.